\documentclass[a4paper,fleqn]{cas-dc}
\usepackage[]{algorithm2e}
\SetKw{Continue}{continue}

\usepackage[authoryear,longnamesfirst]{natbib}

\def\tsc#1{\csdef{#1}{\textsc{\lowercase{#1}}\xspace}}
\tsc{WGM}
\tsc{QE}

\begin{document}
\let\WriteBookmarks\relax
\def\floatpagepagefraction{1}
\def\textpagefraction{.001}

\shorttitle{Retail store customer behavior analysis system: Design and Implementation}    

\shortauthors{Tuan Nguyen Dinh et~al.}  

\title [mode = title]{Retail store customer behavior analysis system: Design and Implementation}

\author[1]{Tuan Dinh Nguyen}[type=editor,
      style=english,
      orcid=0000-0003-4367-6273]
\ead{ndinhtuan15@vnu.edu.vn}

\author[2]{Keisuke Hihara}[type=editor,
      style=english]
\ead{Hihara-k@mail.dnp.co.jp}

\author[1]{Tung Cao Hoang}[type=editor,
      style=english]
\ead{caohoangtung2001@gmail.com }

\author[2]{Yumeka Utada}[type=editor,
      style=english]
\ead{Utada-Y1@mail.dnp.co.jp}

\author[2]{Akihiko Torii}[type=editor,
      style=english]
\ead{Torii-A@mail.dnp.co.jp}

\author[2]{Naoki Izumi}[type=editor,
      style=english]
\ead{Izumi-N@mail.dnp.co.jp}

\author[1]{Nguyen Thanh Thuy}[type=editor,
      style=english]
\ead{nguyenthanhthuy@vnu.edu.vn}

\author[1]{Long Quoc Tran}[type=editor,
      style=english,
      orcid=0000-0002-4115-2890]
      
\cortext[1]{Corresponding author}
\cormark[1]
\ead{tqlong@vnu.edu.vn}
\affiliation[1]{organization={VNU University of Engineering and Technology},
            addressline={144 Xuan Thuy, Cau Giay}, 
            city={Hanoi},
            country={Vietnam}}
            
\affiliation[2]{organization={Dai Nippon Printing Co., Ltd.},
            country={Japan}}

\begin{abstract}
Understanding customer behavior in retail stores plays a crucial role in improving customer satisfaction  by adding personalized value to services. Behavior analysis reveals both general and detailed patterns in  the interaction of customers with a store's items and other people, providing store managers with insight into customer preferences. Several solutions aim to utilize this data by recognizing specific behaviors through statistical visualization. However, current approaches are limited to the analysis of small customer behavior sets, utilizing conventional methods to detect behaviors. They do not use deep learning techniques such as deep neural networks, which are powerful methods in the field of computer vision. Furthermore, these methods provide limited figures when visualizing the behavioral data acquired by the system. In this study, we propose a framework that includes three primary parts: mathematical modeling of customer behaviors, behavior analysis using an efficient deep learning-based system, and individual and group behavior visualization. Each module and the entire system were validated using data from actual situations in a retail store. 
\end{abstract}


\maketitle

\section{Introduction} \label{SECTION:introduction}
The preferences of individuals are visible in their behavior, interactions with other customers or employees, and purchasing activities. Understanding customer behavior in  retail stores is essential in providing a more personal and compelling shopping experience; enhancing store operations; and ultimately improving user experience, sales conversion rates, and revenue. Typically, the  staff at retail stores are relied upon to provide relevant information in most of the studies on client behavior and sentiments. However, when studying a large number of customers, human employees lack the flexibility and reactivity to effectively analyze customer behavior. Consequently, consumer behavior needs to be automatically assessed with minimal delay and tracked over time. \par

\subsection{Related work}
Studies on the prediction of customer behavior in stores is limited in existing literature. Nevertheless, efficient systems can be easily developed. This can be done by mathematically modeling the various gestures made by customers within stores. Moreover, these systems  can be expanded to other gestures for different situations. Furthermore, this mathematical approach simplifies the challenge of decomposing this process into more recognizable sub-problems, such as tracking, detection, and linking them in the system. \par

A few studies have been conducted in this field, such as \cite{popa2010analysis} and \cite{wu2015customer}; however, these systems cannot be generalized to apply to other issues. Moreover, they possess limited inheritance when conducting behavior analysis in settings other than exhibitions or hospitals. In addition, these studies neither structure the modules in a distributed manner, nor provide sufficient empirical analyses of real-world behavior. Furthermore, these systems use conventional machine learning or image-processing methods to recognize relevant behaviors. For example, \cite{popa2010analysis} used the mean shift algorithm for a human tracking module, which is sensitive to complicated backgrounds such as those found in a retail store, and \cite{wu2015customer} used morphological processing and the HOG algorithm to detect people. \par

Recently, owing to advances in deep learning in computer vision, such as \cite{liu2020deep}, \cite{zhao2019object}, \cite{jiao2019survey}, \cite{minaee2021image}, and \cite{zhou2019review}, deep learning models have become more efficient and accurate. Numerous studies on surveillance camera systems have been published, such as tracking \cite{zhang2021fairmot}, \cite{Wojke2017simple} and \cite{Wojke2018deep}, in which algorithms can capture the trajectory of people such as store customers, and detection \cite{bochkovskiy2020yolov4}, \cite{duan2019centernet}, \cite{carion2020end}, in which detection algorithms can use an image as input to create a bounding box around an object such as a human, car, dog, or cat, and determine its location. Additionally, deep learning is extremely effective in recognizing critical client characteristics, such as head poses \cite{yang2019fsa}, \cite{dai2020rankpose}, \cite{ruiz2018fine}. These studies estimated three degrees of angle roll, pitch, and yaw using a face picture clipped by face detection in the preceding phase. These data aid in the determination of the client attention zone in the store using customer behavior systems. A few studies \cite{cao2019openpose}, \cite{toshev2014deeppose}, \cite{sun2019deep}, employed a complex neural network to determine the pose of a human skeleton. The data used to analyze customer behavior are derived from the appearance, gestures, and interactions of customers with other people or objects in the shop. These data are almost entirely gathered via camera images. Today, most stores are equipped with surveillance cameras, which has resulted in the publication of various studies on monitoring customer behavior in-store, including \cite{alfian2020store}, \cite{liu2015customer}, \cite{generosi2018deep}, \cite{yolcu2020deep}, and \cite{liu2018customer}. However, the majority of these publications focus on specific sub-modules of the customer behavior problem and have not yet developed a generic technique or system with a high capacity for module integration. For instance, \cite{yolcu2020deep} focused only on face analysis to ascertain customer interest, \cite{alfian2020store} investigated an approach for determining a customer's browsing behavior, and \cite{liu2018customer} focused on customer pose estimation through a bidirectional recurrent neural network. Additionally, these experiments were conducted mostly in the laboratory and lacked data on actual customer behavior. \par

Customer preference is expressed at the store not only through individual actions such as picking up an item, glancing at the area surrounding the item \cite{liu2017customer}, or approaching this area, but also through group behavior. Group behavior is an efficient way for customers to express their concerns with other objects, such as items or employees. F-formation is a very familiar technique for describing group behavior \cite{pathi2019f}, \cite{kendon1990conducting}, \cite{ciolek1980environment}. In \cite{kendon1990conducting}, the author divides the F-formation group into numerous varieties such as the L-shaped group, the Vis-Vis group, the Circular group, and the Side-by-Side group. In this article, we discuss three different configurations: L-shaped, Vis-Vis, and side-by-side. The first process of determining an f-formation group is group detection, which requires segmenting the crowd into tiny groups. The second phase uses the head pose, body pose, and position of each member to identify the group type. Numerous studies have been published on the initial phase of F-formation \cite{setti2013multi}, \cite{ setti2015f}; nevertheless, these studies assume prior knowledge of the customer's 3D position and face orientation, which are extremely complicated pieces of information in practice. The most recent state-of-the-art study on the F-formation problem is \cite{hedayati2020reform}, which, like previous studies, assumes that the 3D coordinates and face orientation are available from the SALSA benchmark dataset \cite{alameda2015salsa}. In this study, the researchers employed a pipeline structure that comprises three distinct steps: data deconstruction, pairwise classification to construct a correlation matrix for individuals in an image, and reconstruction to cluster individuals in the same group from the correlation matrix; the F-formation module in our study is based on this method. Moreover, we produced F-formation results from the head pose estimation, human pose estimation, and object detection modules to elucidate the connection between the results of these modules and the F-formation result. Due to the scarcity of data on the classification of F-formations in general, we classified them using rules based on the pose and location properties of the members of the group. \par 

\begin{figure*}[ht]
	\centering
	  \includegraphics[width=\textwidth]{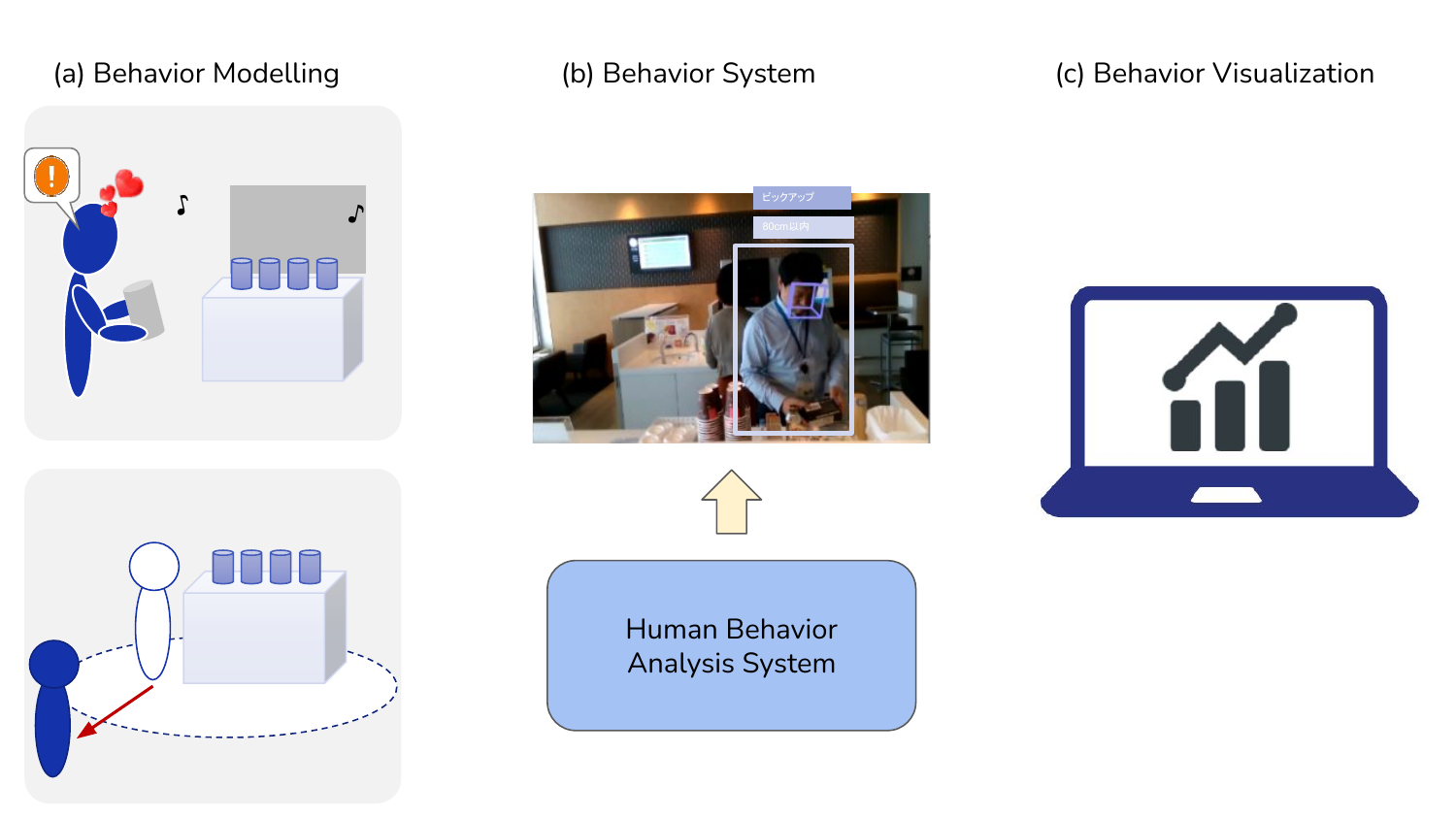}
	\caption{Human Behavior Analysis needs of retail store}
	\label{FIG:overviewproblem}
\end{figure*}

As shown in Fig. \ref{FIG:overviewproblem}, a comprehensive system should comprise three components. Firstly, behavior Modelling, which enables us to present our designs mathematically; secondly, a behavior System, which enables us to use the design from the first part to outline and implement the system; and finally, once the system is implemented, (c) Behavior Visualization, which provides insight into our behavior data. In fact, the current studies on behavior systems only visualize abstract results of behavior data. For instance, \cite{liciotti2014shopper} only describes the average visit time, visitor count, and percentage of interaction, and \cite{liu2017customer} only visualizes trajectories of various movements, such as arm actions. In our study, we describe in detail both personal and group behaviors for a single day during which our system was deployed in a real store.\par

\subsection{Aim of the paper}

This study proposes a comprehensive framework for modelling a behavior analysis system. This provides a better context for designing the system and integrating new modules, or modifying the structure of the system in the future. In-store customer behavior was modelled by considering each individual as a finite-state machine. The system possessed a layered architecture, and its modules were implemented in a distributed approach, which enabled the system to be deployed across various devices. Finally, our methodology was implemented in a real-world retail environment, and behavior data were visualized in detail using both personal behavior and group interaction information.\par

To the best of our knowledge, after a review of the current state-of-the-art models, the primary contributions of our framework are as follows:

\begin{itemize}
    \item Building an approach to modeling customer behavior in the store. With this tool, we can decompose this enormous problem into smaller ones and generalize it to other challenges.
    \item Building a behavior analysis system from modeling. The system can be decentralized to a large number of devices, from which it can optimize speed and leverage the distributed problem's capabilities.
    \item Evaluating the system in-store, where it collects and visualizes data about the behavior of individual and group users. This provides insight into customer behavior.
\end{itemize}

\begin{figure*}[ht]
	\centering
	  \includegraphics[width=\textwidth]{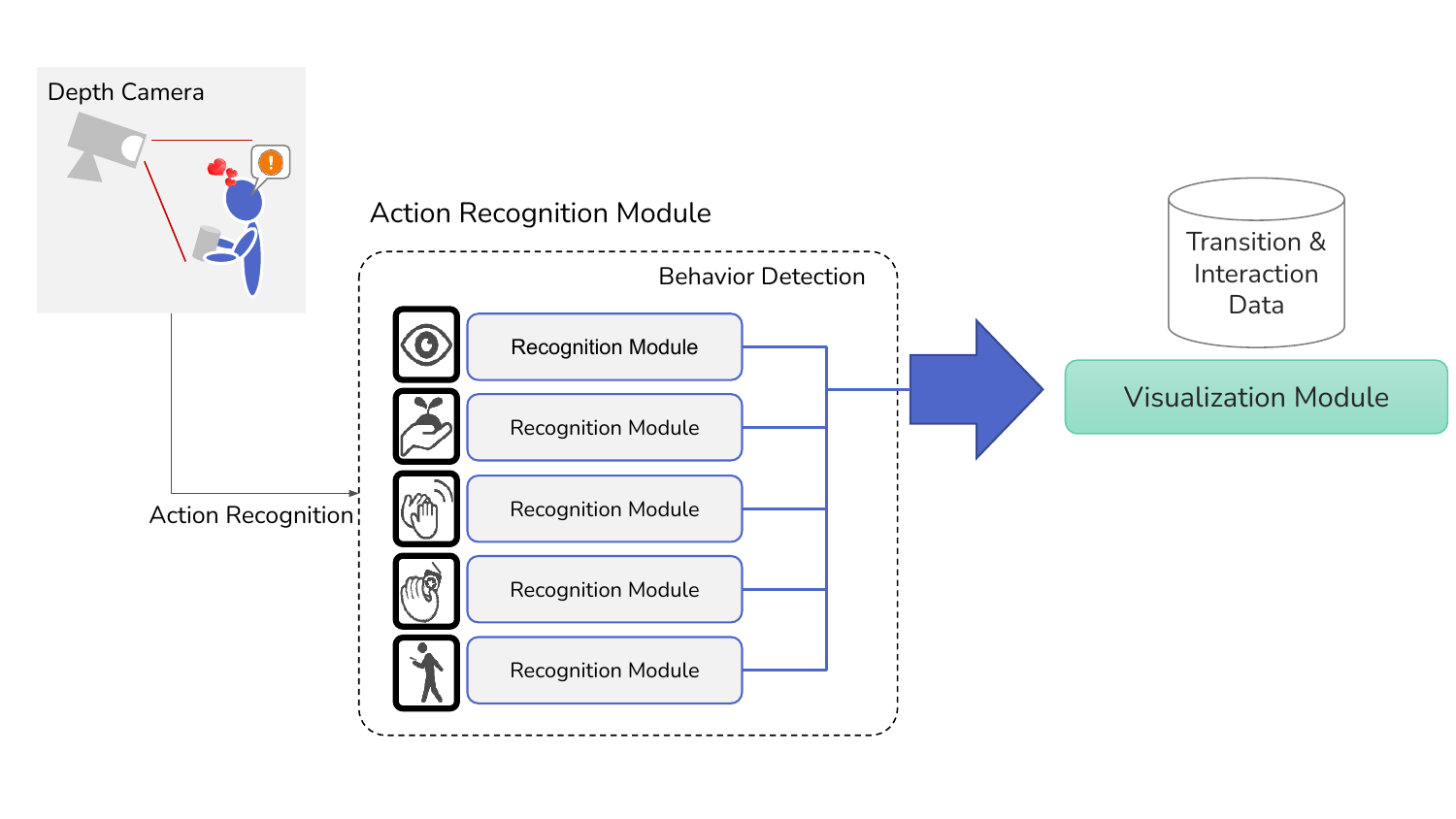}
	\caption{Overview of system architecture}
	\label{FIG:overviewsystem}
\end{figure*}

The rest of the paper is organized as follows: Section \ref{SECTION:sec_system} describes the modeling and design of the system; Section \ref{SECTION:sec_validation} presents the performance analysis for each module in the system, together with the visualization of behavior data; and Section \ref{SECTION:sec_conclude} presents the conclusions.

\section{System} \label{SECTION:sec_system}
In this case, the customer behavior analysis system uses data from sensors, specifically, a camera with depth data. They enable behavior recognition modules to recognize and store data in two forms: transition and interaction data. The transition and interaction data  depicted as components in Fig. \ref{FIG:overviewsystem} represent the general system.

\subsection{Modelling} 

Regarding the problem of customer behavior analysis in the retail business, there are three major questions that the system wishes to address about the customer's activity:\par

\begin{enumerate}
    \item Where do customers go in the store?
    \item Which items pique the customer's interest or attention?
    \item Who do customers interact with during the decision-making process?
\end{enumerate}

According to the answers to the above questions, a customer behavior analysis system should track each person's location and distinguish  them from other customers when they enter the store. Additionally, the system must be aware of the person's field of view or area of interest, as well as what the consumer picks up in the store. If the customer is interested in an item, they must pick it up to inspect it. Moreover, interactions between customers, employees, and other customers are critical in the assessment of  customer behavior, as they provide insight into the consumer's level of interest. Additionally, the system can track the amount of time that employees spend serving consumers. Interestingly, the study \cite{liciotti2014shopper} has developed a system but has not yet developed a model of customer behavior.\par

As can be seen, the analysis of a person's purchasing behavior can be classified into two categories: individual behavior and group behavior. Consequently, this section attempts to mathematically model these two behavior types. \par

\subsubsection{Personal Attribute} \label{peronal_behavior}

The system in the store that represents a human ($H_i$) has the following characteristics that require attention during the purchasing process:

\begin{equation}
H_i=\{\beta_i, id_i, \tau_i, \phi_i, \Lambda_i,o_i\}
\end{equation}

With:
\begin{itemize}
    \item $\beta_i$ represents the three-dimensional coordinates of the system's depth camera and bounding box of $i$th individual.
    \item $id_i$ contains the person's id. The system can follow a person's movements and distinguish them from others in the store using $\beta_i, id_i$.
    \item $\tau_i$ indicates the type of person; it could be a customer or a store staff.
    \item $\phi_i$ is the direction of the person's head, clients typically demonstrate interest through the direction of their heads.
    \item $\Lambda_i$ represents the pose points on the person's arm; the system determines if the person $H_i$ is carrying an object using this property.
    \item $o_i$ is the id of the store item that $H_i$ picked up to view. If an individual $H_i$ does not pick any items, this attribute is set to $null$.
\end{itemize}

By modeling the above individual behavior, the system can identify attributes using methods such as image processing and computer vision. To be precise, the system utilizes the tracking architecture stated in \cite{Wojke2018deep} in conjunction with camera depth data to determine $\beta_i$ and $id_i$ in 3D space with the original coordination from the camera. Using a neural network \cite{sandler2018mobilenetv2} enables the system to classify a person as a customer or employee $\tau_i$ or classify the type of item picked up by the customer $o_i$. The system is based on the study \cite{yang2019fsa} on determining the head pose $\phi_i$ and \cite{cao2019openpose} on estimating the human pose $\Lambda_i$.

\subsubsection{Group Behavior} \label{SECTION:group_behavior}
Individuals in a store can be employees or customers, as shown in the following list:
\begin{equation}
    \mathbb{H}=\{H_i\mid i\in [1,N]\}
\end{equation}

As described in Section \ref{SECTION:introduction}, group identification is a critical module for studying consumer behavior. The system determines the F-formation group behavior using head pose and location data. After identifying the groups, the system classifies them into one of three fundamental types: L-shape, Vis-Vis, and side-by-side, as defined in \cite{pathi2019f}. We assumed that the F-formation group can be created by

\begin{equation}
    \mathbb{G}= f^{\mathbb{G}}(\mathbb{H}; \theta^{\mathbb{G}} = (\theta^{\mathbb{G}}_1, \theta^{\mathbb{G}}_2))
\end{equation}

where $f^{\mathbb{G}}$ is a method that determines and classifies F-formation groups, and the parameter $\theta^{\mathbb{G}} = (\theta^{\mathbb{G}}_1, \theta^{\mathbb{G}}_2)$ specifies two thresholds for the angle effort between each pair in the group to categorize the type of F-formation group. For instance, the system distinguishes between two distinct groups of people.

\begin{equation}
    \mathbb{G}=\{G_1,G_2\}=\{\{H_1,H_3\}, \{H_2,H_4, H_5\}\}
\end{equation}

However, identifying groups of people is a difficult task in consumer behavior analysis. and requires an algorithm to find groups based on F-formation \cite{hedayati2020reform}. \cite{hedayati2020reform} proposes an algorithm that detects F-formation groups based on the distance between individuals and each person's head pose, evaluated based on the SALSA dataset \cite{alameda2015salsa}. Similarly, the system uses the distance between two individuals based on the person's 3D position ($\beta_i$) and head pose ($\phi_i$) to detect F-formation groups. In contrast to \cite{hedayati2020reform}, which relies on distance and head pose annotation data from the SALSA dataset, our method derives these two parameters from sensor data. We discuss the group behavior algorithm in Sections \ref{SECTION:layerbased} and \ref{alg:interact_node}.\par
Additionally, the behavioral system must distinguish between consumers and employees inside the group's behavior, and each person's $\tau_i$ property must be provided.

\subsubsection{State of Customer} \label{SECTION: state_customer}

The previous section outlines the individual attributes and group behavior of customers. However, the attributes of each customer do not define the behavior of each individual. Therefore, this section defines the customer as an object with states, and each state may be considered a behavior. Each client visiting the store is in a variety of states (choose an item, approach item, or leave item), each of which is associated with one or more of the attributes defined in Section \ref{peronal_behavior}. \par

\begin{figure}
	\centering
	  \includegraphics[scale=0.3]{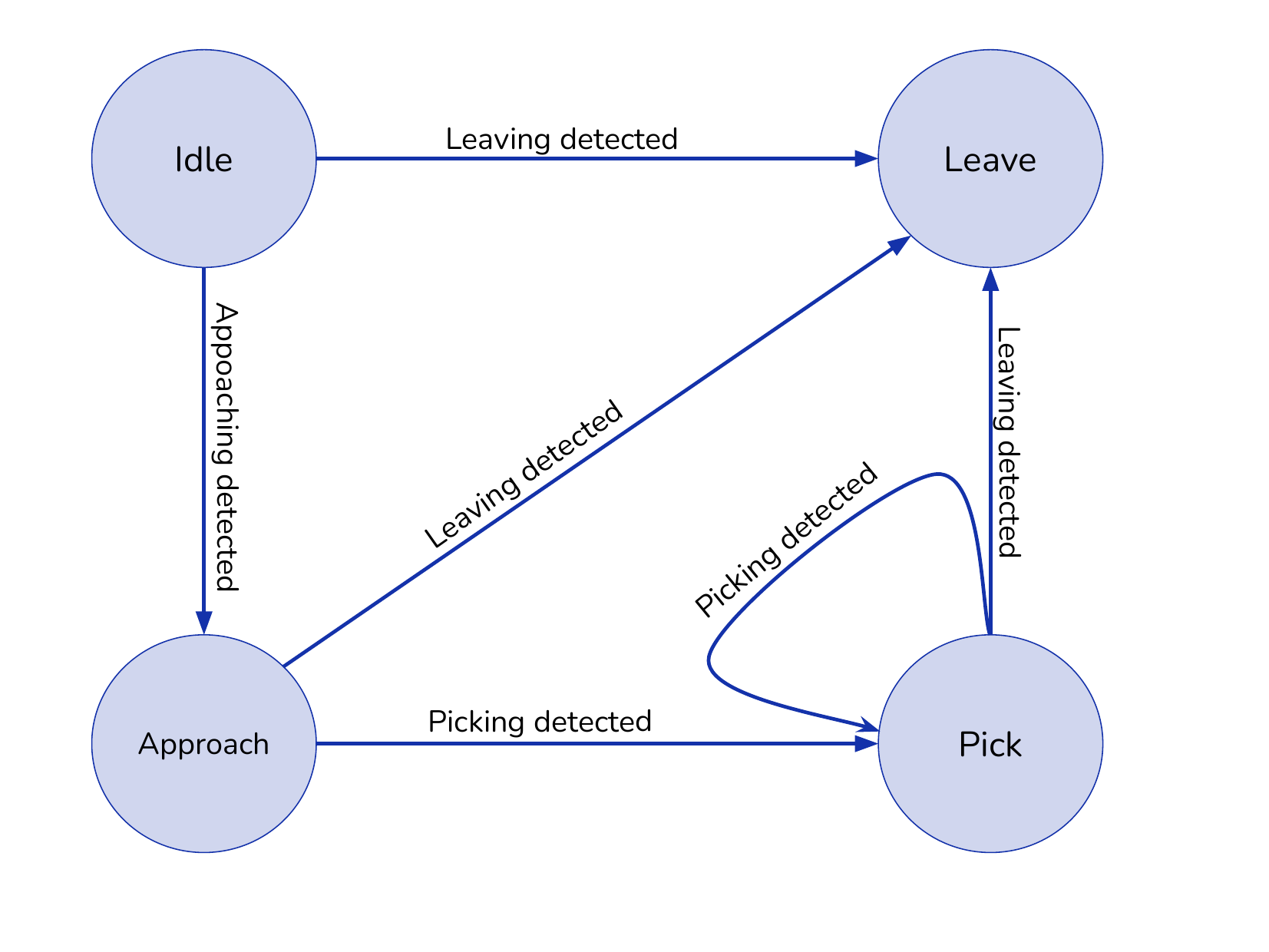}
	\caption{Modeling a person $H_i$ as a finite-state machine with the set of states $Q=\{I, A, L, P\}$. }
	\label{FIG:statemachinemodeling}
\end{figure}

The system considers each person $H_i$ as a finite-state machine, presented by a four-tuple $(S_i^t, Q, q_0, F, f^S)$, where $S_i^t$ is the state of person $H_i$ at time $t$, $Q$ is the set of states of $H_i$, $q_0\in Q$ is the start state of $H_i$, $F\subseteq Q$ is the set of final states of $H_i$, and $f^S$ is the set of transition functions. Fig. \ref{FIG:statemachinemodeling} illustrates state machine modeling for a person $H_i$, with the following $S_i^t$ states:

\begin{itemize}
    \item \textbf{Idle(I)} is the start state assigned to each person when the system detects them. Therefore, we assume that $q_0=\{I\}$,
    \item \textbf{Approach(A)} is the state of the person when the system recognizes that the person is approaching a retail item.
    \item \textbf{Leave(L)} is the state of the person when the system detects that the person leaving the item or suddenly disappears from the frame for a sufficiently long duration. Therefore, we define the set of final states as $F=\{L\}$.
    \item \textbf{Pick(P)} is the state of the person when the system detects that the person is picking up an item.
\end{itemize}

Thus, a person's state is represented as $S_i^t\in Q = \{I,A,L,P\}$. Transforming from state $S_{i}^{t-1}$ to state $S_i^t$ requires several conditional events and depends on the value of $S_{i}^{t-1}$. The set of transition functions $f^S=\{f^{S_A}, f^{S_P}, f^{S_L}\}$ is responsible for transforming the state of $H_i$ as follows:

\begin{itemize}
    \item With $S_{i}^t=I$, is the start state $q_0$ of the person. Thus, it will be set by default when that person appears.
    \item $S_{i}^t=A$ if $S_{i}^{t-1}\in\{I\}$, then the system realizes that person $H_i$ is approaching an item based on the  distance information between the person and item area, using the  transition function $f^{S_A}$.
    \item $S_{i}^t=P$ if $S_{i}^{t-1}\in\{A, P\}$, then the system detects that person $H_i$ is picking up an item via $\Lambda_i$, with transition function $f^{S_P}$.
    \item $S_{i}^t=L$ if $S_{i}^{t-1}\in\{I,A,P\}$, then the system realizes that person $H_i$ is leaving the item area, using the transition function $f^{S_L}$.
\end{itemize}

\begin{figure}
	\centering
	  \includegraphics[scale=0.25]{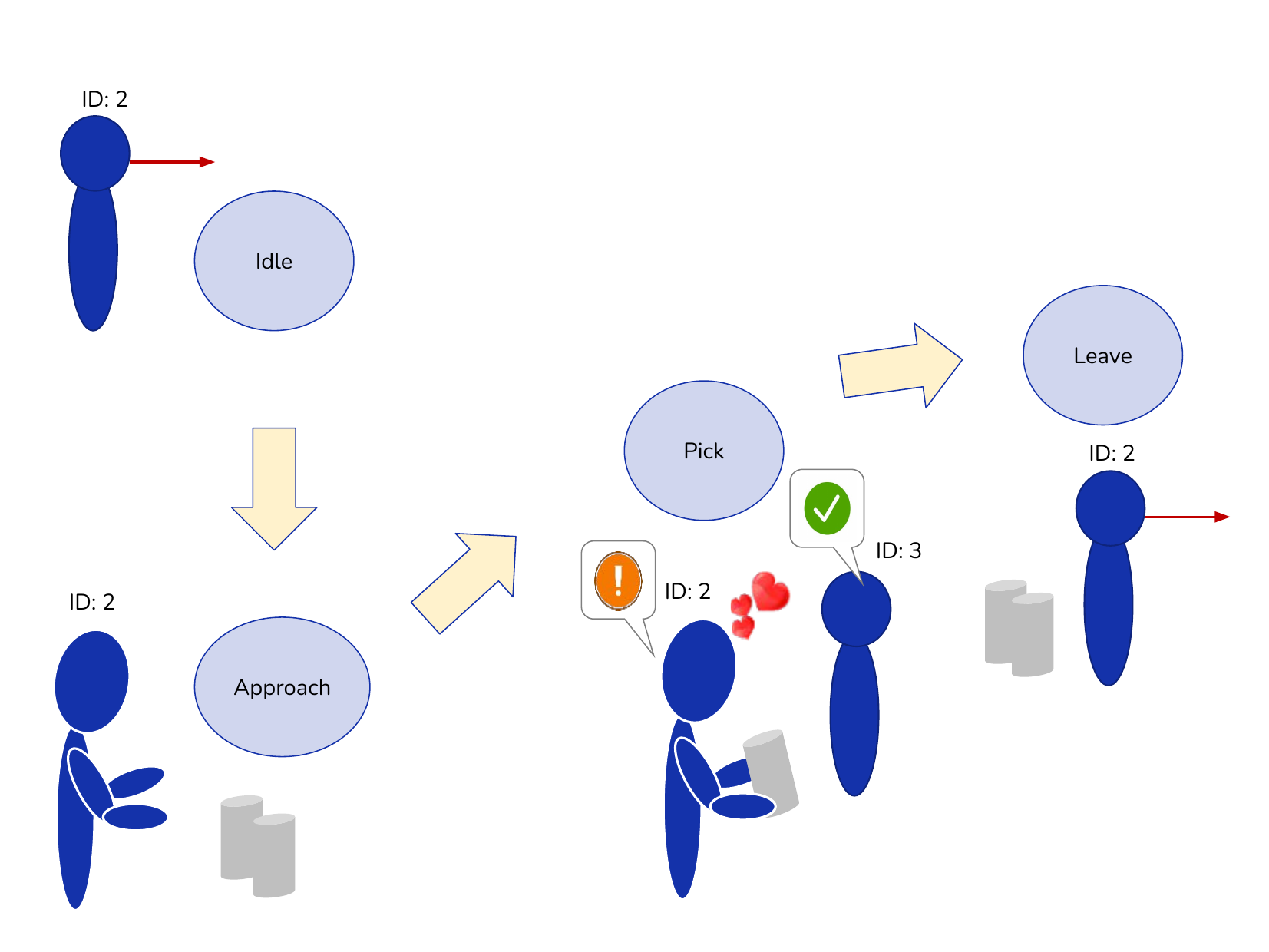}
	\caption{State transition of customer $H_2$: $I\rightarrow A \rightarrow P \rightarrow L$}
	\label{FIG:statetransition}
\end{figure}

Fig. \ref{FIG:statetransition} depicts a state transition of a customer with id $2$ ($H_2$): $I\rightarrow A \rightarrow P \rightarrow L$ and Table. \ref{TAB:statetransition} describes how the state transition is logged back into the database by the system. As indicated in the log, person $H_2$, after being detected by the system, has $I$ state. Subsequently, $H_2$ approaches the item at $05/31/2021, 09:16:31$. The system recognizes that the person has reached the area around the item, and this transition is recorded in the first row of the table. Following this, person $H_2$ is detected to have picked ($P$) an item up at $05/31/2021, 09:16:44$ and moved ($L$) at $05/31/2021, 09:17:30$. Additionally, the system logs the 3D coordinates of the individual throughout each transition based on the information from the depth camera employed by the system. \par

\begin{table*}
\caption{Log for state transition of customer having id 2}\label{TAB:statetransition}
\begin{tabular*}{\tblwidth}{@{} LLLLLLLLL@{} }
\toprule
RowID & PersonID & Prev\_State & State & Distance(m) & Date time & X & Y & Z\\
\midrule
1 & 2 & I & A & 2.3 & 05/31/2021, 09:16:31 & 2.3 & 1.2 & 3.2\\
2 & 2 & A & P & 2.5 & 05/31/2021, 09:16:44 & 2.2 & 1.7 & 3.5\\
3 & 2 & P & L & 4.3 & 05/31/2021, 09:17:30 & 3.4 & 2.4 & 5.8\\
\bottomrule
\end{tabular*}
\end{table*}

\begin{figure*}[ht]
	\centering
	  \includegraphics[width=\textwidth]{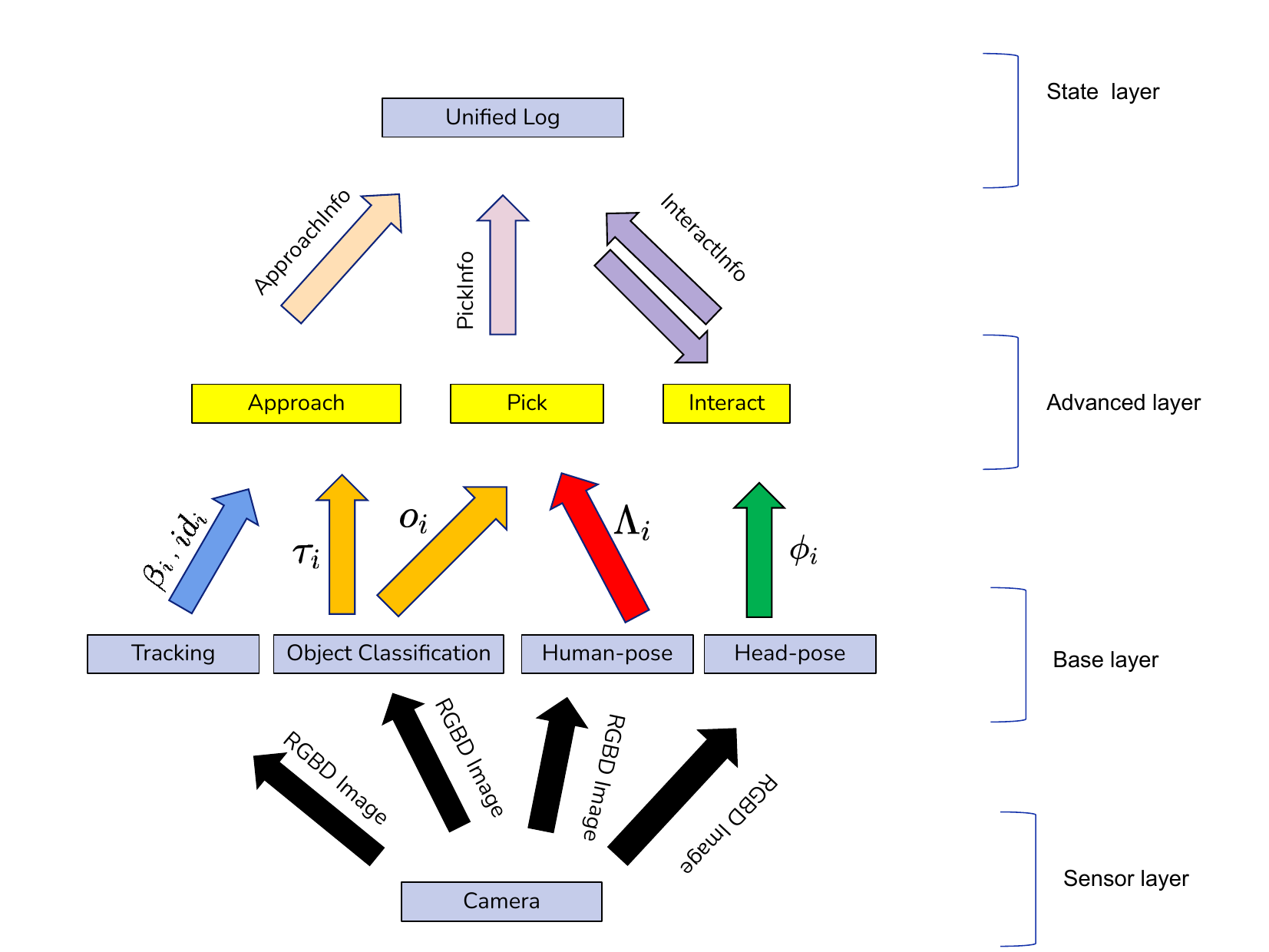}
	\caption{Layer-based system architecture}
	\label{FIG:layerbasedsystem}
\end{figure*}

\subsection{System Design and Implementation}

In the previous section, a person's attributes were represented by the small actions a person makes when he or she enters the store, the state of a person, or the behavior of a group of people. These are high-level actions based on attributes. Therefore, an efficient hierarchy of behaviors and attributes is required for interaction between higher-order behaviors and basic attributes.\par

\subsubsection{Layer-based system} \label{SECTION:layerbased}

The system is organized into layers, as illustrated in Fig.\ref{FIG:layerbasedsystem}, and comprises four layers:

\begin{itemize}
    \item \textbf{Sensor layer} is a layer that works with sensor devices, specifically our system using cameras with depth data, RGBD image (Intel Realsense D435 \cite{tadic2019application}). In addition to the camera, the system can be easily expanded to include additional sensors, such as an acoustic sensor or a multisensor system.
    \item \textbf{Base layer} contains modules to identify a person's attributes, for example the module \textit{Tracking}, \cite{Wojke2018deep} provides the system with information about the attribute bounding box ($\beta_i$ ) and id ($id_i$). Modules \textit{Human-pose} \cite{cao2019openpose}, \textit{Object classification}, and \textit{ head-pose } \cite{yang2019fsa} provide information about $\Lambda_i$, $o_i$, $\tau_i$, and $\phi_i$.
    \item \textbf{Advanced layer} combines the attribute information in \textbf{Base layer} related to certain states and sends it to \textbf{State layer}. The module \textit{Approaching} in this layer is responsible for aggregating the results of the \textit{Tracking} and \textit{Object Classification} modules at the \textbf{ base layer} and post-processing these results before transferring the location and type of person to the \textbf{State layer} through a message called \textit{ApproachInfo}. Similarly, the module \textit{Pick} aggregates attribute information from \textit{Human-pose} and \textit{Object Classification} to send information about who is picking up items and what items are being picked up through a message called \textit{PickInfo}. \ textit{Interact}aggregates information from \ textit{Object Classification}and \ textit{head-pose} and submits information about groups and the $id_i$ of the individuals in that group through a message named \textit{InteractInfo}. In particular, the \textit{Interact} module receives the $\beta_i, id_i$ information from a person's \textit{ unified log }. Messages \textit{ApproachInfo}, \textit{PickInfo}, and \textit{InteractInfo} are shown in Fig. \ref{FIG:advanced_message}.
    \item \textbf{State layer} is supported by the \textbf{Advanced layer} and determines which state $H_i$ ($S_i^t$) the tracked individual is currently in. Moreover, this layer controls the state transition of all individuals when the detection system is in operation. This layer also logs the state transitions of people in the store and their behavior, of which there are two types: individual and group behavior.
\end{itemize}

Table. \ref{TAB:module_deep} lists modules that recognize system attributes in \textbf{Base layer}, these modules all use deep learning techniques. \par

The proposed architecture of the system, which is divided into layers and subdivided into behavioral recognition modules, enables the modules to replace algorithms efficiently. For example, in \textit{Tracking}, we can replace the tracking algorithm with various algorithms such as deepsort \cite{Wojke2018deep} and \cite{zhang2021fairmot} without changing the architecture of the entire system and affecting other modules. \par

\begin{table*}
\caption{Method for each attribute recognition module}\label{TAB:module_deep}
\begin{tabular*}{\tblwidth}{@{} LLL@{} }
\toprule
Module & Method & Using pretrained\\
\midrule
Tracking & Deepsort \cite{Wojke2018deep} & Yes\\
Human pose estimation & Open pose \cite{cao2019openpose} & Yes\\
Head pose estimation & FSA-net \cite{yang2019fsa} & Yes \\
Store-staff classification & Mobilenet \cite{sandler2018mobilenetv2} & No\\
Item classification & Mobilenet \cite{sandler2018mobilenetv2} & No\\
\bottomrule
\end{tabular*}
\end{table*}

For a person $H_i$, the \textbf{ sensor layer} and \textbf{ base layer} enable the system to compute basic human attributes $\{\beta_i,$ $id_i, \tau_i, \phi_i, \Lambda_i,o_i\}$. We assume that the system determines the state $S_i^t$ based on the following attributes: 

\begin{equation}
    S_i^t=f^S(\beta_i, id_i, \tau_i, \phi_i, \Lambda_i,o_i, S_{i}^{t-1}; \theta^S)
\end{equation}

where $f^S$ denotes the set of transition functions for identifying the current state $S_i^t$ and $\theta^S$ denotes the parameter of the method. \par

A state is associated with only a subset of the properties of person $H_i$. More precisely, with $S_i^t=A$: 

\begin{equation}
    S_i^t=f^{S_{A}}(\beta_i,id_i, S_{i}^{t-1}; \theta^S=(\theta^S_1, \theta^S_2, \theta^S_3, \theta^S_4))
\end{equation}

where $f^{S_{A}}$ is the transition function that determines whether the current state of person $S_i^t = A$ with condition $S_{i}^{t-1} \in\{I\}$. We assume that person $H_i$ is identified as approaching the item area if $H_i$ approaches the item in both three-dimensional and two-dimensional space, based on $\beta_i$. In three-dimensional space, a person $H_i$ is considered to approach an item if his/her distance to the item area is sufficiently small several times over a specified duration. Owing to the instability of distance estimates, two-dimensional space information should be utilized if a person's bounding box overlaps with the area surrounding the item several times over a specific duration. Consequently, we suppose that $\theta^S_1$ is the window size specifying the duration to verify personal information that satisfies the 2D and 3D conditions; $\theta^S_2$ is the distance threshold in the 3D condition, $\theta^S_3$ is the threshold for the number of events required before a person satisfies the 3D condition; and $\theta^S_4$ is the threshold for the number of events required before a person satisfies the 2D condition. $f^{S_{A}}$, is described in detail in Algorithm \ref{alg:approach_node}. All parameters were selected via a grid search on a validation set.\par

With $S_i^t = P$,

\begin{equation}
    S_i^t=f^{S_P}(\Lambda_i, o_i, S_{i}^{t-1}; \theta^S=(\theta^S_5, \theta^S_6))
\end{equation}

where $f^{S_P}$ is the transition function used to determine whether the current state of an individual is $S_i^t=P$. To detect the picking activity and classify objects, we employed a voting algorithm. to be precise, based on $\Lambda_i$, we can recognize $H_i$ picking up an item. Then, the algorithm crops the bounding box around the hand to classify the type of item in the hand. The algorithm repeats the procedure and votes if the picking action is recognized as larger than $\theta^S_5$, confirming state $S_i^t=P$. Similarly, the classification model samples and classifies items $\theta^S_6$ times and returns the id with most occurrences. $f^{S_P}$ is described in detail in Algorithm \ref{alg:pick_node}, \par

With $S_i^t = L$,

\begin{equation}
    S_i^t=f^{S_L}(\Lambda_i,o_i, S_{i}^{t}; \theta^S=(\theta^S_7, \theta^S_8, \theta^S_9, \theta^S_{10}))
\end{equation}

Similar to the algorithm for identifying the approaching state, the transition function $f^{S_L}$ uses four parameters to determine a person's state, $S_i^t=L$, with condition $S_{i}^{t-1} \in {I,A,P}$. The algorithm is based on both two-dimensional and three-dimensional information to detect the state of departure. In three dimensions, person $H_i$ is considered to leave an item if their distance is large enough several times over a specified duration. In two dimensions, if a person's bounding box does not overlap with the area surrounding the item over a certain time period, they are considered to be leaving the item.\par

For the modeling of a group of people, as was described in Section. \ref{SECTION:group_behavior}, there are three types of groups. To construct a group of people we need three attributes $\beta_i, id_i, \phi_i,\tau_i$, so the function of the \textit{Interact} module in \textbf {Advanced layer} has form:

\begin{equation}
    \mathbb{G} = f^{\mathbb{G}}(\{\beta_i,id_i, \phi_i,\tau_i \mid i \in [1, N]\}, \theta^{\mathbb{G}} = (\theta^{\mathbb{G}}_1, \theta^{\mathbb{G}}_2))
\end{equation}

where $N$ is number of people being detected by the system. \par

The two modules \textit{Approach} and \textit{Pick} support the \textbf{State layer} to manage the state of person $H_i$ in the system, whereas the \textit{Interact} module supports this layer to manage the actions of group behavior : L-shape, Vis-Vis, and side-by-side.

\subsubsection{Message-based process}\par

The message-based approach \cite{ozansoy2007real}, \cite{o2014gentle} enables a process to be a publisher or subscriber to communicate with others via messages carrying information that the process wants to deliver. These messages can be transmitted using various protocols \cite{quigley2009ros}, simplifying the implementation of the system across different devices. \par

\begin{figure*}
	\centering
	  \includegraphics[width=\textwidth]{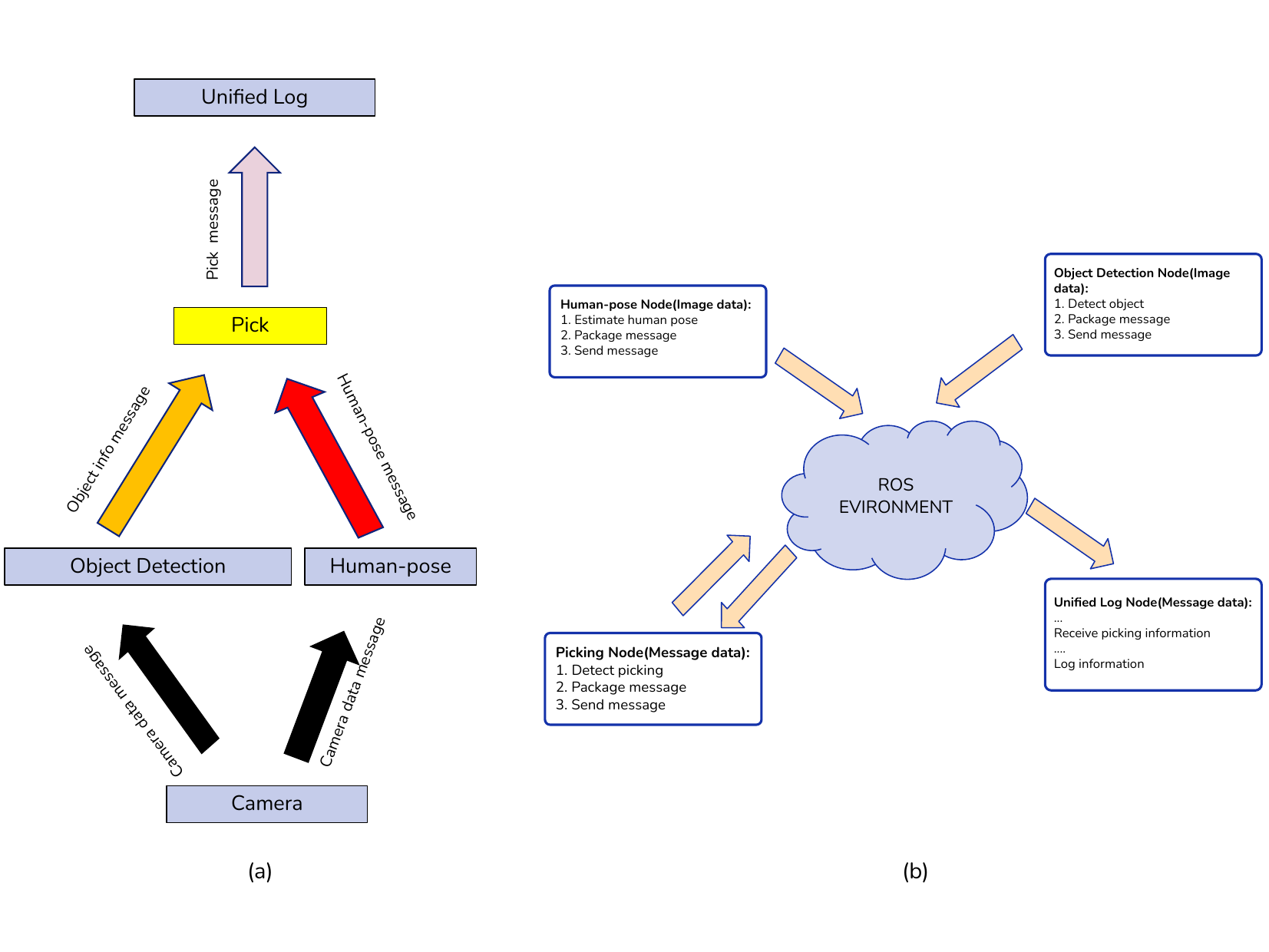}
	\caption{(a) Description of system work following Message-based process scheme. (b) Details of each process in ROS environment.}
	\label{FIG:algorithmpicking}
\end{figure*}

\begin{figure}
	\centering
	  \includegraphics[scale=0.3]{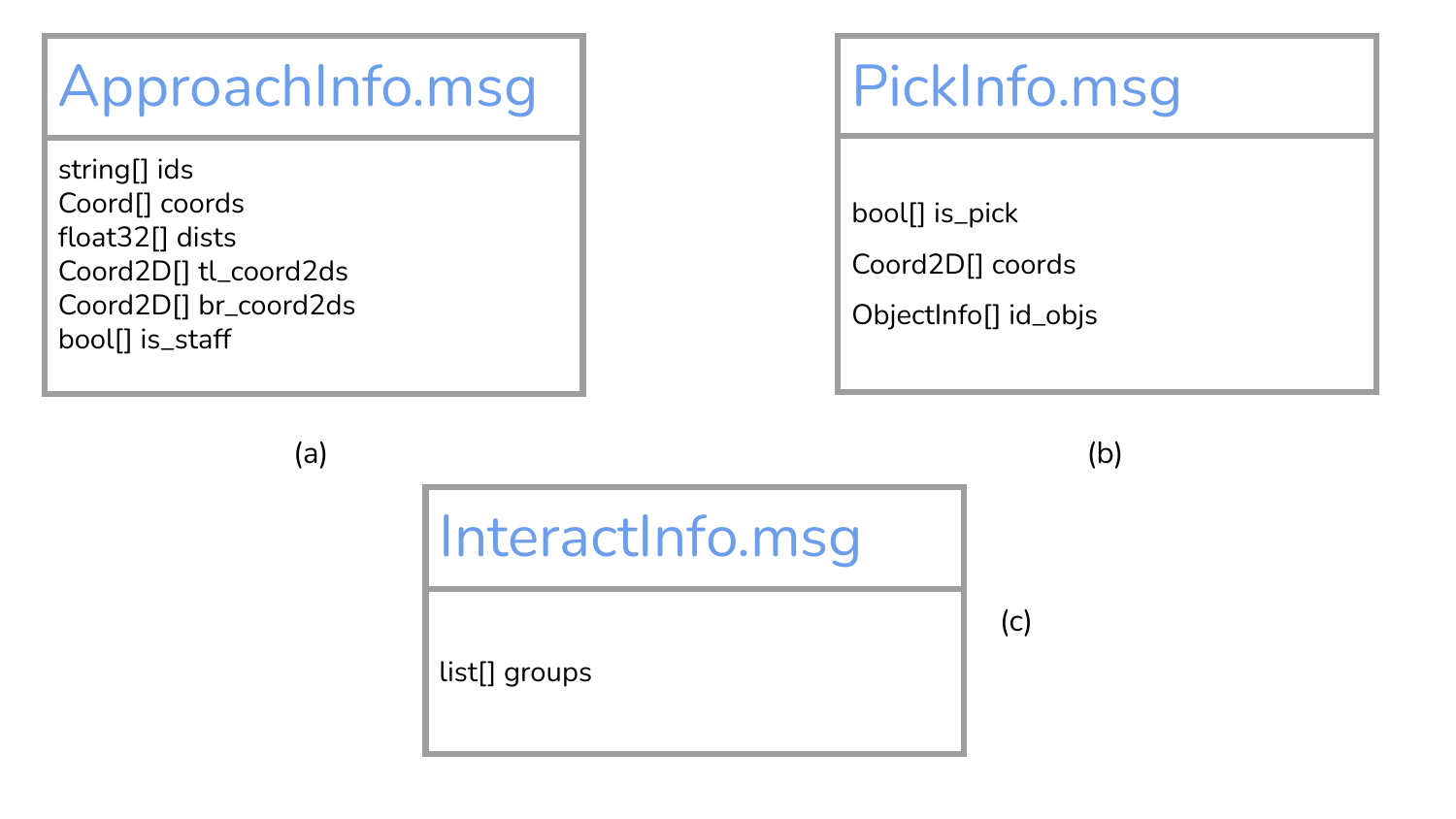}
	\caption{Message definition for information of \textit{Approach}, \textit{Pick} and \textit{Interact} node}
	\label{FIG:advanced_message}
\end{figure}

Figure \ref{FIG:algorithmpicking} (a) describes the data flow to each module of the system for the purpose of determining the state $P$ of each customer in the store. Each module within the system is referred to as a node. The camera node obtains data from the camera, compresses it, and passes it to \textit{Human-pose} and \textit{Object Detection} nodes. Both nodes \textit{Human-pose} and \textit{Object Detection} are subscribers that receive messages from the management node \textit{Camera} and also publishers that send information to the \textit{Pick} node at a higher level. Similarly, the \textit{Pick} node receives messages from the two related nodes in the lower layer and forwards them to the \textit{Unified Log} node. Figure \ref{FIG:algorithmpicking} (b) describes the actual system at deployment time divided into processes. Each rectangle represents an algorithm that is implemented in a process and communicates with each other through the ROS environment \cite{mishra2018ros}, \cite{seib2016ros}. Figures \ref{FIG:advanced_message} illustrates message definition for \textit{Approach}, \textit{Pick} and \textit{Interact} node, which are used to communicate with other nodes. \par

\begin{algorithm}
    \caption{Main algorithm $f^{S_A}$ detecting \textit{Approach} state}\label{alg:approach_node}
    
    \KwData{$\mathbb{H}=[H_1,\dots,H_n]; \theta^S_1, \theta^S_2, \theta^S_3, \theta^S_4$}
    \KwResult{Verify if $S_i^t=A$ for each $H_i$}
    
    \For{$H_i \gets H_1$ \KwTo $H_n$}{
        
        \If{$S_i^t \neq I$}{
            \Continue
        }
        
        \If{length of personal data of $H_i \leq  \theta^S_1$}{
            \Continue
        }
        
        $T_{2d}=0$\\
        $T_{3d}=0$
        
        \For{$h_i$ in the newest $\theta^S_1$ information of $H_i$}{
            \If{distance between $h_i$ and item $\leq \theta^S_2$}{
                $T_{3d} = T_{3d} + 1$
            }
            \If{$h_i$ overlap item area}{
                $T_{2d} = T_{2d} + 1$
            }
        } 
        \If{$T_{3d}\geq \theta^S_3$ and $T_{2d}\geq \theta^S_4$}{
            $S_i^t=A$
        }
    }
\end{algorithm}

\begin{algorithm}
    \caption{Main algorithm $f^{S_P}$ detecting \textit{Pick} state}\label{alg:pick_node}
    
    \KwData{$\mathbb{H}=[H_1,\dots,H_n]; \theta^S_5, \theta^S_6$}
    \KwResult{Verify if $S_i^t=P$ for each $H_i$}
    
    \For{$i \gets 1$ \KwTo $n$}{
        $T_i = 0$ // threshold picking time \\
        $L_{i} = []$ // list consisting of classified voting items.
    }
        
    \For{$H_i \gets H_1$ \KwTo $H_n$}{
        
        \If{$S_i^t \neq \{A,P\}$}{
            \Continue
        }
        
        $h_i =$ the newest information of $H_i$ \\
        \If{$h_i$ is detected picking by $\Lambda_i$}{
            $T_i = T_p + 1$ \\
            $bbox = $ image cropped around the hand \\
            $o = $ id of an item classified by model \\
            
            \If{$o$ is not null}{
                $L_{i}.\text{append}(o)$
            }
        }
        \Else{
            $T_i = 0$ \\
            $L_{i} = []$
        }
        
        \If{$T_i\geq \theta^S_5$ and $len(L_i)\geq \theta^S_6$}{
            $S_i^t=P$ \\
            $o_i$ = id with most occurrences in $L_i$,
        }
    }
\end{algorithm}


\begin{algorithm}
    \caption{Main algorithm $f^{S_{\mathbb{G}}}$ detecting F-formation group}
    \label{alg:interact_node}
    
    \KwData{$\mathbb{H}=[H_1,\dots,H_n]; \theta^{\mathbb{G}}_1, \theta^{\mathbb{G}}_2$}
    \KwResult{Type $T_i$ for each group $G_i$}
    
    Deconstruct human information \\
    Classify pairwise \\
    $\{G_1,\dots,G_n\} = $ Reconstruct F-formation group \\
    
    \For{$G_i \gets G_1$ \KwTo $G_n$ }{
        
        \If{len($G_i$) $>2$ }{
            $T_i$ = "Circular"
        }
        \Else{
            $\delta = $ effort angle of the two people\\
            \If{$\theta^{\mathbb{G}}_1 \leq \delta \leq \theta^{\mathbb{G}}_2$}{
                $T_i = $ "L-shape"
            }
            \If{$\delta < \theta^{\mathbb{G}}_1$}{
                $T_i = $ "Side-by-Side"
            }
            \If{$\delta > \theta^{\mathbb{G}}_2$}{
                $T_i = $ "Vis-Vis"
            }
        }
    }
    
\end{algorithm}

Algorithms \ref{alg:approach_node} and \ref{alg:pick_node} are used to determine the $A and P$ states, respectively. The two algorithms receive data containing personal information $\mathbb{H}=[H_1,\dots,H_n]$ from \textbf{Base layer}. These data also include history information of each $H_i$. Thus, we refer to $h_i$ as the history record of a person $H_i$. \textit{Unified Log} receives this information from the lower layer and switches the status for each person analyzed by the system according to the rules defined in Section \ref{SECTION: state_customer}. This node then logs information about the state transitions for each person, and groups are generated every second. The algorithm employed in node \textit{Interact} was derived from \cite{hedayati2020reform}. Algorithm \ref{alg:interact_node} was used for detecting F-formation groups, which uses the three processes of \cite{hedayati2020reform} to cluster the crowd into small groups before classifying them.\par

\section{Validation and Behavior Visualization} \label{SECTION:sec_validation}

This section describes the quantitative evaluations conducted to evaluate the ability to recognize a person's state and behavior of a group in a store. The system was installed in a Vietnamese phone retail store, and customer behavior  at a table showing four key store items was analyzed. The accuracy of modules that identify personal and group behaviors was evaluated during store operating hours from 9 a.m. to 10 p.m., as detailed in Section \ref{SECTION:module_validation}. Sections \ref{SECTION:personal_viz} and \ref{SECTION:group_viz} visualize the statistical data for a single day at the store according to personal and group behavior. Three devices were used to implement the system: a realsense D435 camera, embedded computer Nvidia Jetson Nano, and a PC equipped with an Nvidia 1080Ti card, in which the Jetson Nano ran a node that published camera data acquired from the store. The PC was placed in a room for aesthetic reasons. The system is easily scalable to multiple modules, runs on a broad range of devices, and can be installed in any store. The validation set created using data from another day enables us to search for parameters; in this case, we find the best parameter set described in Table \ref{TAB:chosenparameter}.


\begin{table*}
\caption{Selected value via grid search on validation day data}\label{TAB:chosenparameter}
\begin{tabular*}{\tblwidth}{@{} LLLLLLLLLLLLLLLL @{} }
\toprule
Parameter & $\theta^S_1$ & $\theta^S_2$ & $\theta^S_3$ & $\theta^S_4$ & $\theta^S_5$ & $\theta^S_6$ & $\theta^S_7$ & $\theta^S_8$ & $\theta^S_9$ & $\theta^S_{10}$ & $\theta^{\mathbb{G}}_1$ & $\theta^{\mathbb{G}}_2$ \\
Chosen value & 7 & 1.8 & 4 & 5 & 8 & 5 & 5 & 4 & 5 & 4 & $\pi/3$ & $2\pi/3$ \\
\bottomrule
\end{tabular*}
\end{table*}

\subsection{State validation} \label{SECTION:module_validation}

The objective of this section is to assess the system's predictive ability for individual and group behaviors. The metrics used are the number of samples true positive ($TP$), false positive ($FP$), false negative ($FN$), and
$$
Precision = \frac{TP}{TP+FP};
Recall = \frac{TP}{TP+FN}.
$$

\begin{table*}
\caption{Personal state evaluation}\label{TAB:state_evaluation}
\begin{tabular*}{\tblwidth}{@{} LLLLLL@{} }
\toprule
State & TP & FP & FN & Precision & Recall\\
\midrule
Approach(A) & 117 & 47 & 60 & 0.71 & 0.66\\
Leave(L) & 1759 & 123 & 10 & 0.93 & 0.995\\
Pick(P) & 32 & 35 & 14 & 0.71 & 0.52\\
\bottomrule
\end{tabular*}
\end{table*}

Table. \ref{TAB:state_evaluation} quantifies the accuracy of the system predicting the states of persons in the store, ignoring the default $I$ state assigned to a person when first detected by the system. The number of samples of the $L$ state is highest, with the number of samples of $TP, FP, FN$ being $1759, 123, 10$, since all states $I, A, P$ can move to state $L$. In contrast, the $P$ state appears least frequently because there is only a limited possibility that a large number of people will pick up a product from the area the system analyzes in a day. Numerous individuals approached but did not pick up an item. Consequently, the algorithm for detecting the $I, L, P$ states was built using data from a day other than evaluation day. \par

\begin{table}
\caption{Store-staff classification evaluation, using Mobilenet\cite{sandler2018mobilenetv2}}\label{TAB:customer_classification}
\begin{tabular*}{\tblwidth}{@{} LLLLLL@{} }
\toprule
TP & FP & FN & Precision & Recall\\
\midrule
2665 & 357 & 460 & 0.88 & 0.85\\
\bottomrule
\end{tabular*}
\end{table}

Table \ref{TAB:customer_classification} describes the accuracy of the store's customer and employee classification function in the \textit{Object Classification} module. This module uses the MobileNet model \cite{sandler2018mobilenetv2} and uses $253480$ for a human image data sample of which $67749$ is a customer sample. The training model had an accuracy of $98.15\%$ when working on the validation set.  Table \ref{TAB:customer_classification} presents the accuracy of this module tested on a different date. \par

\begin{table}
\caption{F-formation group recognition evaluation}\label{TAB:group_detection}

\begin{tabular*}{\tblwidth}{@{} LLLLLL@{} }
\toprule
TP & FP & FN & Precision & Recall\\
\midrule
11752 & 11598 & 2588 & 0.5 & 0.82\\
\bottomrule
\end{tabular*}
\end{table}

With the group identification and F-formation classification module, the system uses \cite{hedayati2020reform} to detect groups and classify them based on the $\phi$ and $\beta$ of individuals at any given time. The performance of the module is described in Table \ref{TAB:group_detection}. It exhibited a precision of $0.5$, recall of $0.82$, for $FP$ samples of $11598$ groups. \par

\subsection{Personal Behavior Analysis Visualization} \label{SECTION:personal_viz}

This section presents the outcomes of the system's logging of state and human qualities during operation at a store. \par
\begin{figure*}
	\centering
	\includegraphics[width=\textwidth]{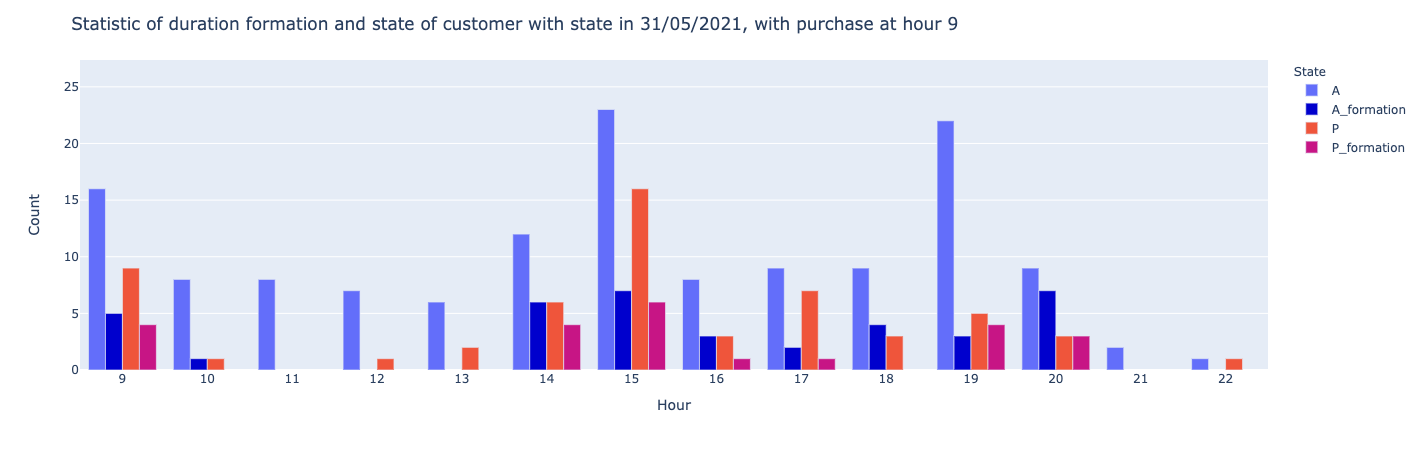}
	\caption{Statistic for state $A$ and $P$ for people in store with and without forming group (F-formation) condition}
	\label{FIG:vizindividual1}
\end{figure*}

It can be seen that the number of customers approaching and the number of customers picking up the product are two states associated with the purchase. In this section, the statistics about the states $A$ and $P$ are only analyzed on a single customer via $\tau$. The graphs describe the figures for these two states in terms of count or duration with and without the formation of an F-formation group. Figure \ref{FIG:vizindividual1} depicts the number of states $A$ and $P$ generated each hour (the number of customers approaching the product area and the number of customers picking up the product) from 9 a.m. to 10 p.m., including statistics. This figure also provides state statistics when the customer interacts with a group. For instance, at 9 a.m., when the actual purchase occurs, more than $15$ state approaching item ($A$) is performed, and nearly $10$ pick item up actions occur. When the customer is a member of the group, the number of activities for these two states is $5$ and $4$. From the graph, it can be seen that when the number of state $A$ increases, the number of the state $P$ also increases, reaching a peak of 15 at p.m. The corresponding peaks of $A$ and $P$ are $22$ and $17$, respectively.\par

\begin{figure*}
\centering
\includegraphics[width=\textwidth]{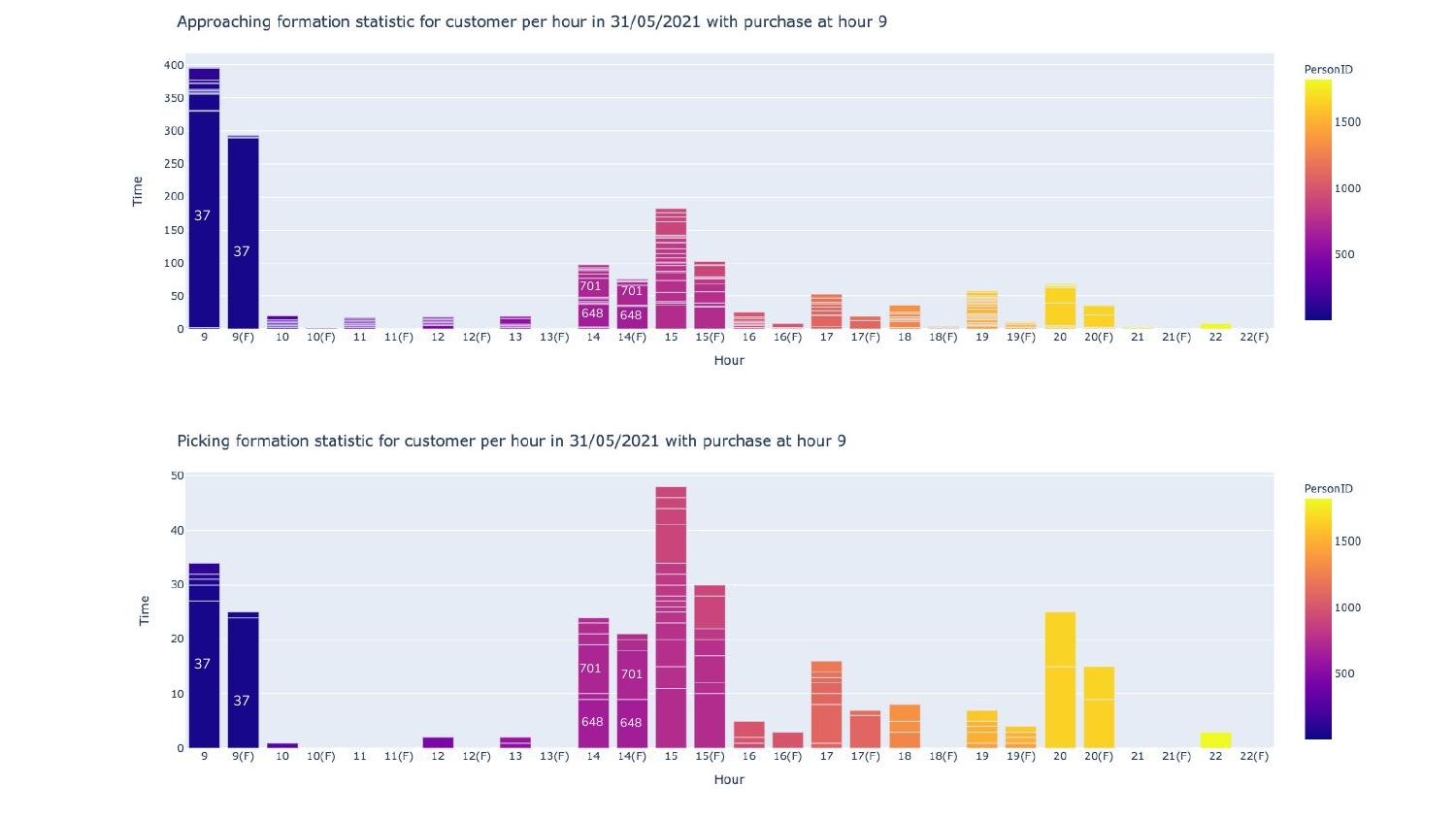}
\caption{Statistic duration of states $A$ and $P$ for individuals in the store with and without group formation (F-formation) condition}
\label{FIG:vizindividual2}
\end{figure*}

Figure \ref{FIG:vizindividual2} describes the duration of state $A$ and state $P$ for one hour, according to personal identifier $id_i$, the duration of a person's state $A$ is the time that elapsed from the person's approach to the product to their  departure from the product, in seconds. Duration of state $P$ is the number of seconds that the person takes to pick up an item. For example, Fig. \ref{FIG:vizindividual2} shows that at 9 a.m.,  customers stood next to the product for a total of 400 s, 300 s of which was the amount of time that the customers stood in a group. Customer with $id_i=37$ stood next to the product for the longest amount of time, which was approximately $340$ s. Similarly, the image in Fig. \ref{FIG:vizindividual2} shows that customer with $id_i=37$ took the longest time to pick up the product at 9 a.m. with an interval of nearly 29 seconds, of which approximately 24 seconds were spent interacting in a group. It is similar for the $id_i=648, 701$ at 14.pm. \par

\begin{figure*}
	\centering
	  \includegraphics[width=\textwidth]{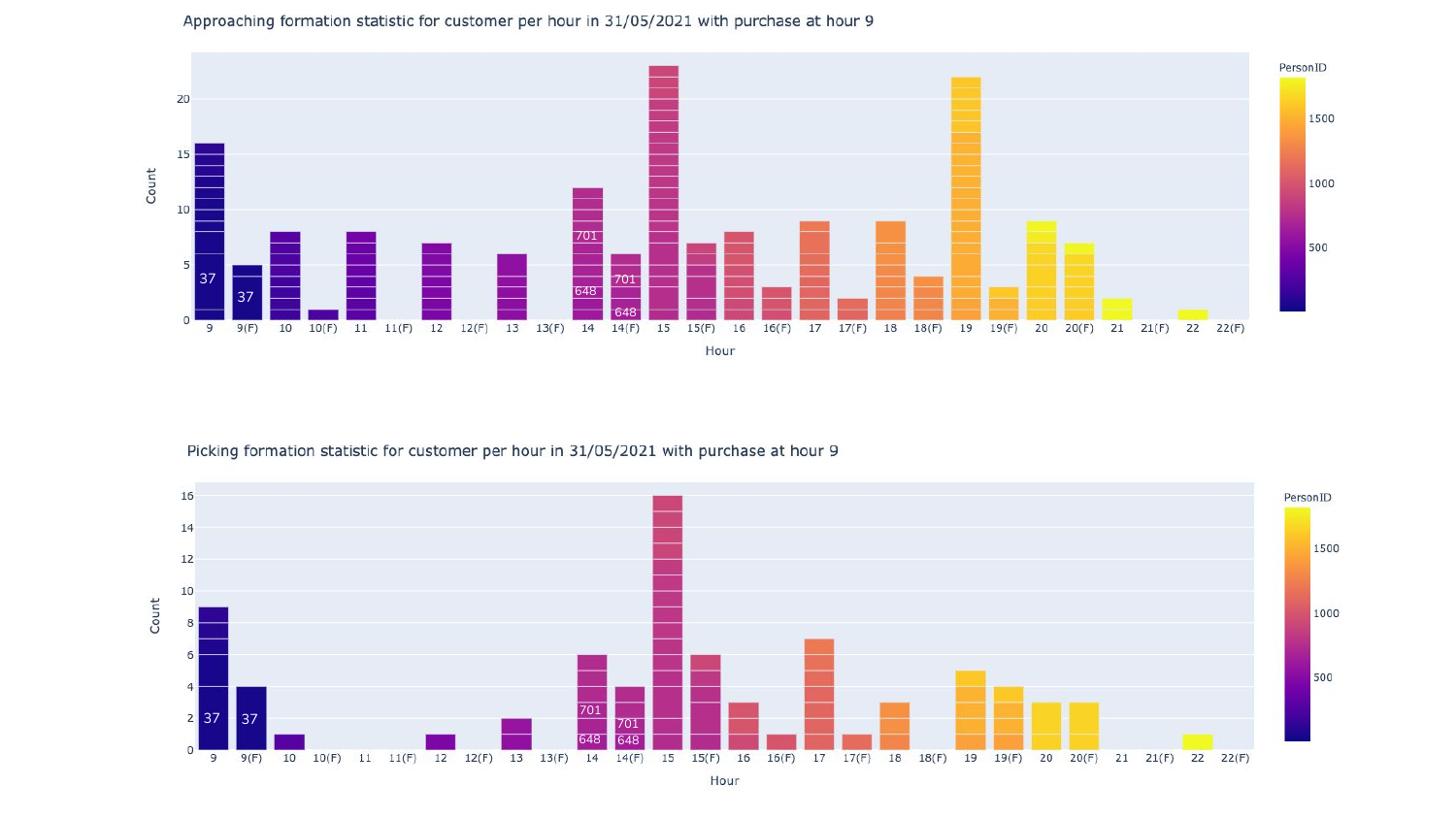}
	\caption{Statistics of states $A$ and $P$ for individuals in store with and without group formation (F-formation) condition}
	\label{FIG:vizindividual3}
\end{figure*}

Figure \ref{FIG:vizindividual3} presents the number of times customers approach the item and the number of times they pick up the item in an hour. For example, at 9 a.m., the person with $id=37$ makes $4$ approaches the item area, then picks up an item 4 times to generate a total of $4$ $P$ states, for which the number remains unchanged even if the person with $id=37$ joins a group. It is similar for $id=701$ and $id=648$ at 14pm. \par

\begin{figure*}
\centering
\includegraphics[width=\textwidth]{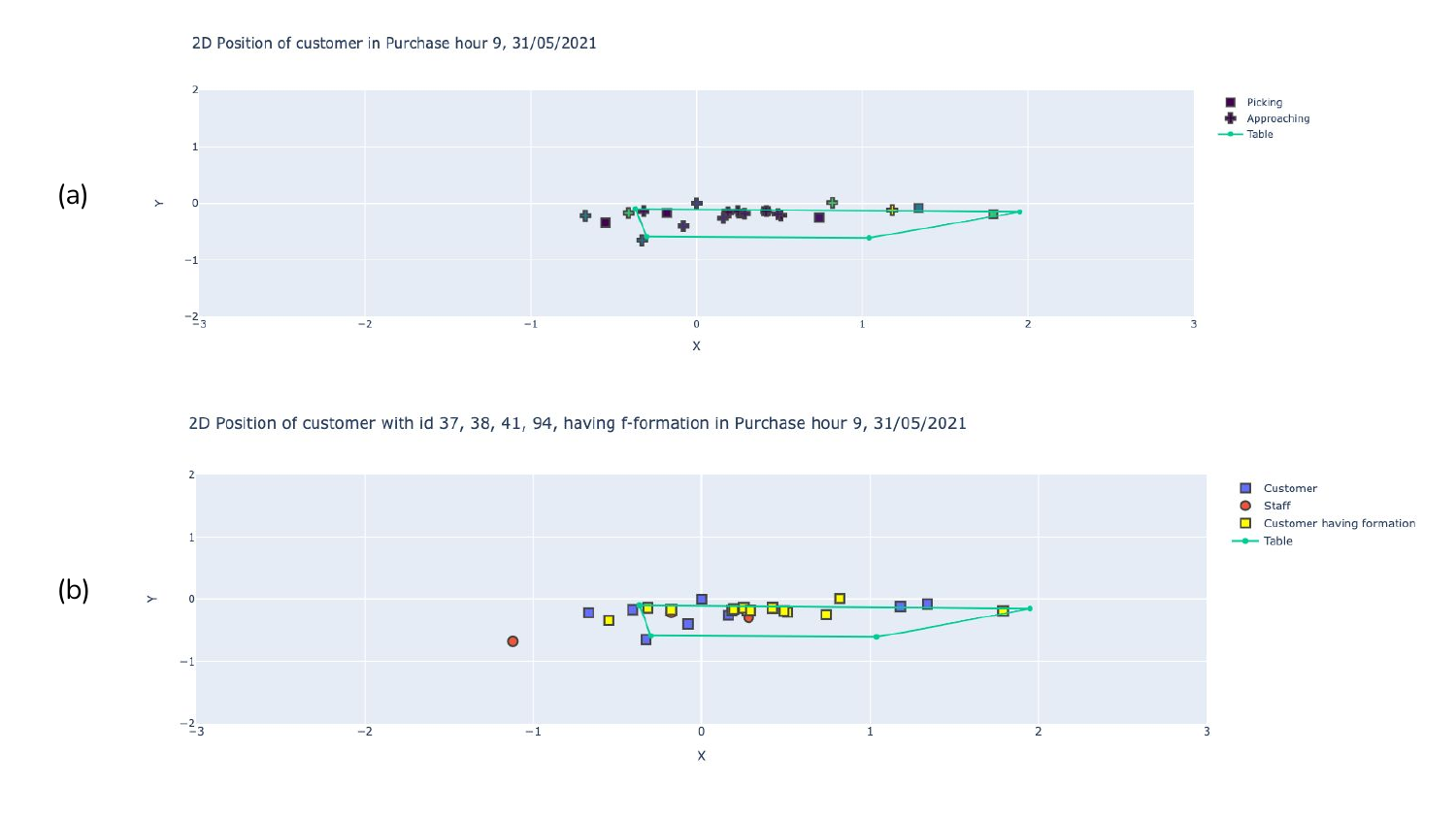}
\caption{(a) 2D position of customer state during purchase hour (9 a.m.) on 5/21/2021. (b) 2D position of customer with IDs 37, 38, 41, 94 with and without forming groups; the red dot represents staff position.}
\label{FIG:vizindividual4}
\end{figure*}

Figure \ref{FIG:vizindividual4} shows the two-dimensional coordination of customers and employees under f-formation conditions during the hour when the purchase was made while the system was operating. To be precise, Fig. \ref{FIG:vizindividual4} (a) depicts the locations at which the customer states occurred during purchase hour (9 a.m.), \ref{FIG:vizindividual4} (b) depicts the location of the customer state, while the customer was interacting in a group using a yellow point. the figure also plots the location of the store's employee (red point). The green rectangle represents the table on which the products were placed. It defines the area that is used to determine the state $A$ of the customers when they enter this area.\par

\subsection{Group Behavior Analysis Visualization} \label{SECTION:group_viz}

This section aims to calculate the time spent by customers interacting with employees in the store (customer–staff). The Customer-staff group includes at least one customer and one staff member. In addition, this section visualizes the group interaction statistics along with the time taken by customers in approaching ($A$) and picking up the product ($P$).\par

\begin{figure*}
	\centering
	  \includegraphics[width=\textwidth]{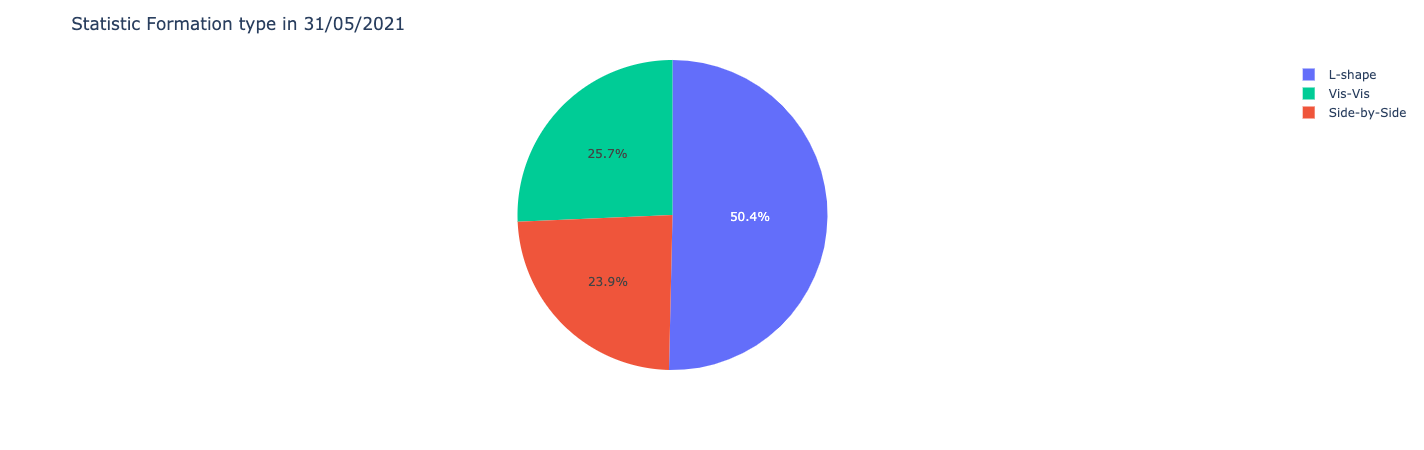}
	\caption{Statistic for F-formation group types in one day.}
	\label{FIG:vizgroup1}
\end{figure*}

In the previous section, three types of F-formation groups were introduced: L-shaped, Vis-Vis, and side-by-side. Figure \ref{FIG:vizgroup1} lists the number of instances of these three formations in a single day when the system was deployed in the store. The number of L-shaped, Side-by-Side and Vis-Vis groups, were $50.4\%$, $23.9\%, and 25.7\%$, respectively. \par

\begin{figure*}
\centering
\includegraphics[width=\textwidth]{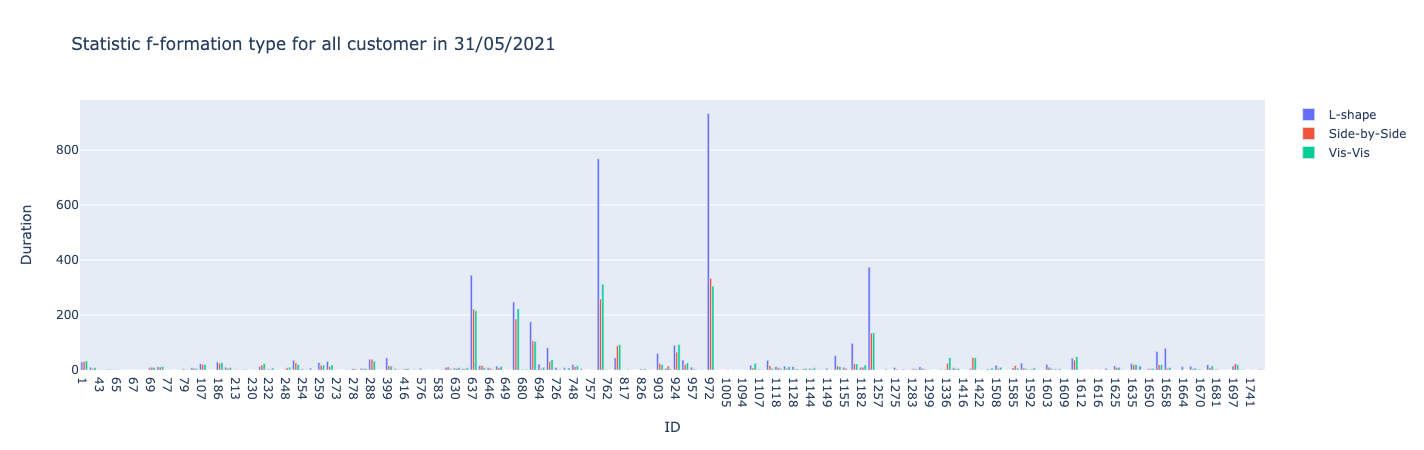}
\caption{Statistics for time spent (in seconds) in each F-formation group type for all customers in one day. \textit{Note: Due to an excessive number of customers, some $id_i$ are not shown.}}
\label{FIG:vizgroup2}
\end{figure*}

\begin{figure*}
\centering
\includegraphics[width=\textwidth]{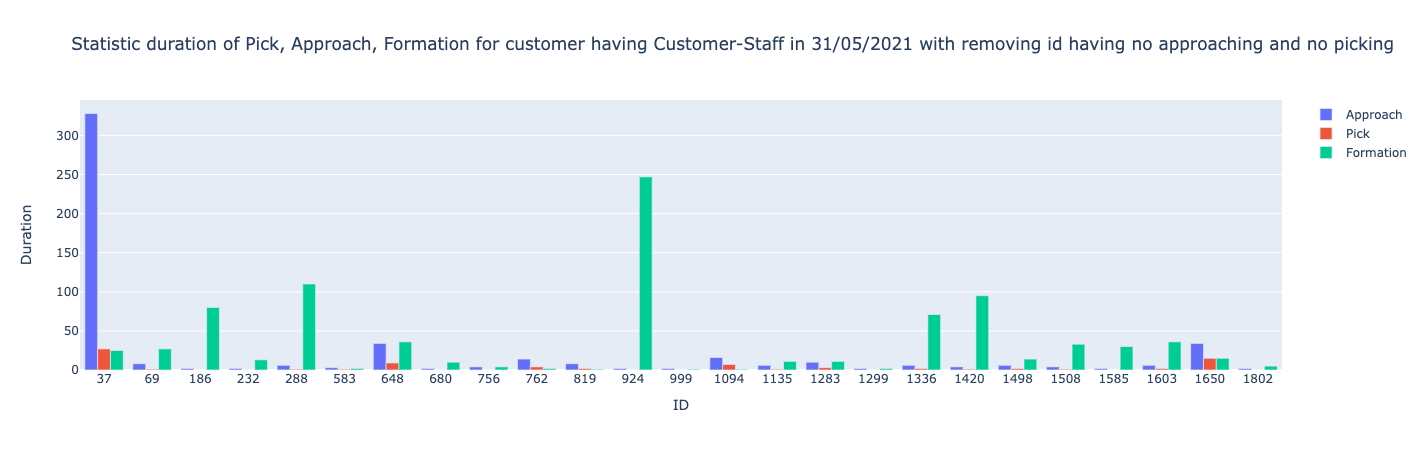}
\caption{Statistics for time spent (in seconds) in the Pick  and Approach states and in standing in F-formation for all customers with the condition that the customer has executed either the approach or pick up state}
\label{FIG:vizgroup3}
\end{figure*}

\begin{figure*}
\centering
\includegraphics[width=\textwidth]{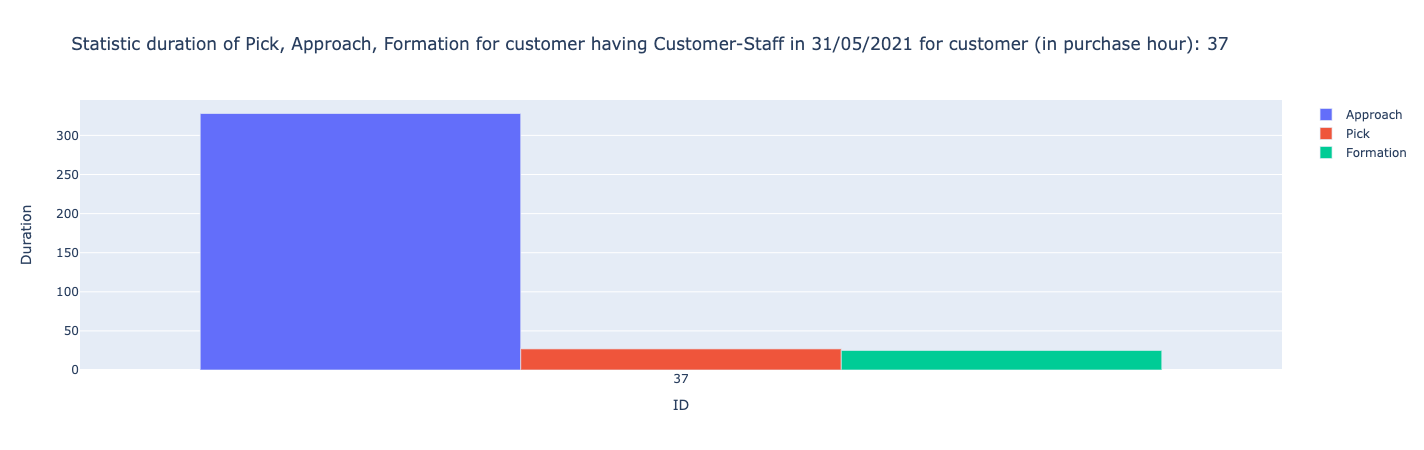}
\caption{Statistics for time spent (in seconds) standing in Customer-Staff group for all customers forming F-formation group.}
\label{FIG:vizgroup4}
\end{figure*}

Figure \ref{FIG:vizgroup2} presents the statistics for the amount of time each customer spent in the various F-formation group types. It can be seen that the customer with $id=972$ had the longest overall interaction time in the group, with approximately $1000$s spent in the L-shaped group, while the combined time spent in the remaining two types of groups are less than $400$s. In Fig. \ref{FIG:vizgroup3}, data on the duration of the customer states are shown, with the condition that each customer  executed both states $A$ and $P$. Customer with $id=37$ took the longest time, $300$s, to approach.  In contrast, picking up items and interacting in a group took an equal amount of time, which was approximately $20$s. Figure \ref{FIG:vizgroup4} presents the statistics of the time spent by customers interacting with store staff during the hour of purchase. In this case, only customers with $id=37$ satisfy this condition. \par

\section{Concluding remarks}
\label{SECTION:sec_conclude}
{ In this study, we proposed a framework for analyzing customer behavior including modeling of customers with purchase-relevant attributes, the system design for the modeling and evaluation in the practical store}. Based on these attributes, customer states ($I, A, P,$ and $L$) were introduced to make customer management more efficient in the system. Based on these states, and their transition in and out of them, each customer in the system was considered a finite-state machine. The transitions from one state to another were assigned certain constraints to ensure that the system did not assign states erroneously to the customers. A four-layer structure was recommended to efficiently organize customer attributes and states, and message-based processing was employed to incorporate customer modeling into the system. {Experiments conducted in an actual store demonstrate that our suggested system can efficiently recognize behaviors.  We evaluate each primitive module in the \textbf{Base layer} to  \textbf{State layer}, which provides us with performance evaluation in all modules in the system.} {Modeling customer behavior allows us to utilize strong mathematical frameworks and expand to other complex behaviors. Furthermore, we could conveniently integrate new behavior recognition modules into our system.} {In this research, we conducted many experiments and visualizations about individual and group behaviors at the practical store. Through these visualizations, the store owners could have insight into their customers. Our system recognizes massive behaviors such as Approaching, Picking, Leaving and attributes such as pose and tracking identification.} {Because of privacy, we cannot retrieve customer identification to identify which behavior is related directly to purchase action. In the future, we expect that we could have the identification information of customers in the experiment to research the factor or the chain of behavior related to purchase. Furthermore, we also want to apply Dynamic Bayesian Network to our modeling and system to capture uncertainties and inaccuracy factors.}
\section*{Acknowledgments}
The authors would like to thank the VNU University of Engineering and Technology, Dai Nippon Printing Co., Ltd., for providing financial support for this study. 

\bibliographystyle{cas-model2-names}

\bibliography{cas-refs}


\end{document}